\documentclass[letterpaper, 10 pt, conference]{ieeeconf}  % Comment this line out if you need a4paper

\IEEEoverridecommandlockouts                              % This command is only needed if
\usepackage{times}
\usepackage{epsfig}
\usepackage{graphicx}
\usepackage{amsmath}
\usepackage{amssymb}
\usepackage{subcaption}
\usepackage{multirow} % Required for multirows
\usepackage{colortbl}
\usepackage{hyperref}

\title{\LARGE \bf
	Improving Human Annotation in Single Object Tracking
}

\author{Yu Pang$^{1}$, Xinyi Li$^{1}$, Lin Yuan$^{2}$ and Haibin Ling$^{3}$ % <-this % stops a space
	\thanks{*This work was not supported by any organization}% <-this % stops a space
	\thanks{$^{1}$Yu Pang and Xinyi Li are with Department of Computer and Information Sciences, Temple University, Philadelphia, PA, USA
		{\tt\small \{yu.pang, xinyi.li\}@temple.edu}}
	\thanks{$^{2}$Lin Yuan is with Amazon Web Services, Palo Alto, CA, USA
		{\tt\small lnyuan@amazon.com}}
	\thanks{$^{3}$Haibin Ling is with Department of Computer Science, Stony Brook University, Stony Brook, NY, USA
		{\tt\small hling@cs.stonybrook.edu}}%
}

\begin{document}

\maketitle
\thispagestyle{empty}
\pagestyle{empty}

%%%%%%%%% ABSTRACT
\begin{abstract}
   Human annotation is always considered as ground truth in video object tracking tasks. It is used in both training and evaluation purposes. Thus, ensuring its high quality is an important task for the success of trackers and evaluations between them. In this paper, we give a qualitative and quantitative analysis of the existing human annotations. We show that human annotation tends to be non-smooth and is prone to partial visibility and deformation. We propose a smoothing trajectory strategy with the ability to handle moving scenes. We use a two-step adaptive image alignment algorithm to find the canonical view of the video sequence. We then use different techniques to smooth the trajectories at certain degree. Once we convert back to the original image coordination, we can compare with the human annotation. With the experimental results, we can get more consistent trajectories. At a certain degree, it can also slightly improve the trained model. If go beyond a certain threshold, the smoothing error will start eating up the benefit. Overall, our method could help extrapolate the missing annotation frames or identify and correct human annotation outliers as well as help improve the training data quality.
\end{abstract}

%%%%%%%%% BODY TEXT
\section{Introduction}

Visual object tracking is one of the key tasks in robotics. It's widely used in SLAM~\cite{fuentes2015visual}, Visual Servoing~\cite{kragic2002survey}, Unmanned Aerial/Ground/Underwater Vehicle Navigation~\cite{bonin2008visual}~\cite{lategahn2011visual}, Augmented Reality~\cite{billinghurst2015survey}, Planar Object Tracking~\cite{liang2018planar}~\cite{wang2018constrained}, Human motion tracking~\cite{moeslund2001survey} etc. Tracking task is normally served as the core algorithm to achieve the goal. Besides the conventional template based tracking, most tracking algorithms (trackers) are learning based, especially as deep learning becomes popular, it often can be found in emerging trackers.

To ensure the high quality of the learning based trackers, ground truth is critical to solve the problem. We need reliable ground truth for both training and evaluation. In visual tracking scenario, collecting training data is a non-trivial task. There is no natural task that incentivizes people to label the object over the timespan of a whole video clip. Hence we need dedicated work on creating ground truth dataset for visual tracking cases. Over the past decade, the community has come up with many valuable datasets, including Visual Tracker Benchmark~\cite{WuLimYang13}, VOT dataset~\cite{VOT_TPAMI}, MOT dataset~\cite{MOTChallenge2015} etc. Some crowd sourcing platforms like Amazon Mechanical Turk, Amazon SageMaker etc. also contributes a lot in completing large volume of annotations. Unlike many other tasks, annotating objects in similar frames can be very boring and easily results in inaccurate bounding boxes. Sometimes due to the deformation or occlusion, the different assumptions lead to inconsistent annotations.

In this paper, we visualize this human annotation result issues and give some analysis. We then propose our method to help correcting these issues by registering the images and smoothing the trajectory. It has several applications, for example, it can be used as a guidance to detect outlier annotations and help correct them. With a certain amount of corrections, it can also help improving the training data quality and improve the models. It can extrapolate the missing annotation frames, this would speed up the annotation process.

\section{Related Work}

In the past decade, the community has made great progress in getting more and more annotated datasets. Especially since the deep learning proves its power, more training data is required to achieve better performances. Heng F. et al.~\cite{fan2019lasot} has categorized the tracking datasets into two buckets: dense benchmarks and sparse/(semi-)automatic benchmarks. Dense benchmarks provide bounding boxes in every frame or every other frame to capture the fine granularity of the trajectories. Examples include OTB~\cite{WuLimYang13}, TC-128~\cite{liang2015encoding}, VOT~\cite{VOT_TPAMI}, NUS-PRO~\cite{li2015nus}, UAV~\cite{mueller2016benchmark}, NfS~\cite{kiani2017need}, Got-10K~\cite{huang2018got} etc. While in the sparse benchmarks, the human labelers annotate objects every a few frames, e.g. 5 to 30 frames. Examples are ALOV++~\cite{smeulders2013visual}, ImageNet Video~\cite{russakovsky2015imagenet}, YT-BB~\cite{real2017youtube} and OxUvA~\cite{valmadre2018long}. Further more, there are other datasets that use tracker outputs as annotations to fill in the missing frames by human labelers, e.g. TrackingNet~\cite{muller2018trackingnet} proposes to use DCF tracker~\cite{henriques2014high} as the annotation filler in between each 1 fps human labeled ground truth. In this way, it achieves a pseudo-dense annotated dataset.

Finding the correspondence in two images has been extensively studied as image registration problem~\cite{zitova2003image}, object tracking~\cite{yilmaz2006object}, motion estimation~\cite{jakubowski2013block}, stereo matching~\cite{hamzah2016literature} and video stabilization~\cite{amisha2015survey} etc. Although the purpose from each scenario is different, the core idea is to find the transformation between two images that can be read from the same coordination system. A set of methods aim at finding the stable feature points at pixel level in two images, e.g. SIFT~\cite{lowe1999object}, SURF~\cite{bay2006surf}, so that image alignment algorithms like RANSAC~\cite{fischler1981random} can be applied to estimate the transformation. Another set of methods try to warp one image into the other, so that the pixel level difference is minimized, e.g. ECC~\cite{evangelidis2008parametric} etc. In terms of a video sequence, to stably mapping them into the same coordination system, there are more recent update strategies that can be applied, e.g. Robust staged RANSAC Tracking~\cite{dutta2014video}, L1-optimal camera paths~\cite{grundmann2011auto}.

In our work, we focus on correcting the errors in human annotations instead of generalizing it to annotate the frames for us like in TrackingNet. The assumption is different in that we are suspicious when human annotation is inaccurate. While in TrackingNet, the human annotation is considered as ground truth and other frames are extrapolated. We acknowledge the high precision of human annotations and the limitation of our method, so we only try to correct them if the discrepancy is large. We use a combined strategy of the image alignment techniques, so that it fits in the video sequence scenario instead of only two images.

This paper is broken down as follows: in Sec.~\ref{sec:eval}, we demonstrate our observations and problems in human annotations. we explain our methodology in Sec.~\ref{sec:proposed_method}. Then following up with experiments to illustrate how it is useful in different scenarios in Sec.~\ref{sec:experiment}. We also discussed the limitations in Sec.~\ref{sec:discussion}.

\section{Annotation Quality Evaluation}
\label{sec:eval}
\subsection{Human Annotation Quality Observations}
\label{sec:human_annotation}
\begin{figure}
	\begin{subfigure}[t]{.22\textwidth}
		\centering
		\includegraphics[width=\linewidth]{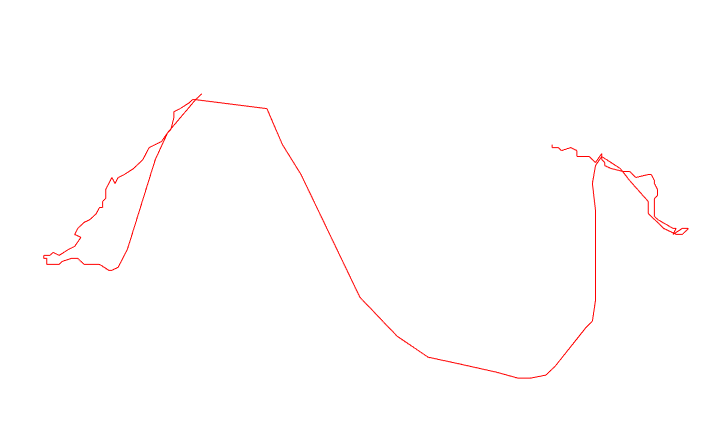}
		\caption{TB-50 Biker}
	\end{subfigure}
	\hfill
	\begin{subfigure}[t]{.22\textwidth}
		\centering
		\includegraphics[width=\linewidth]{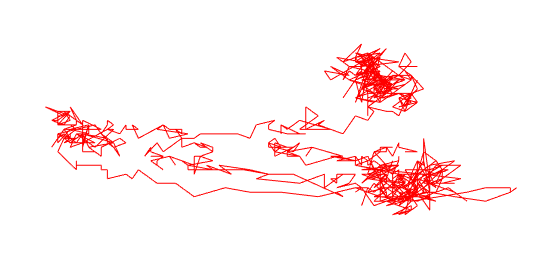}
		\caption{TB-100 FaceOcc2}
	\end{subfigure}

	\medskip

	\begin{subfigure}[t]{.22\textwidth}
		\centering
		\includegraphics[width=\linewidth]{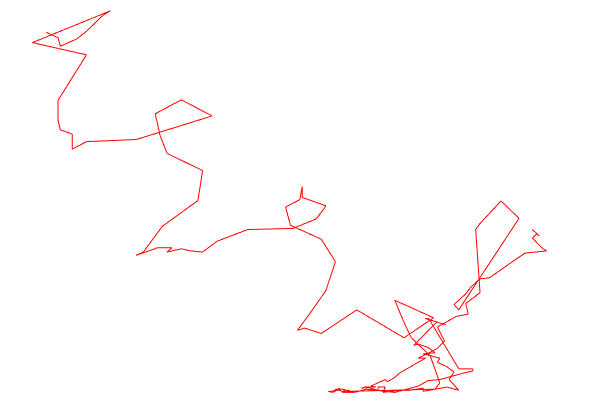}
		\caption{VOT2015 Butterfly}
	\end{subfigure}
	\hfill
	\begin{subfigure}[t]{.22\textwidth}
		\centering
		\includegraphics[width=\linewidth]{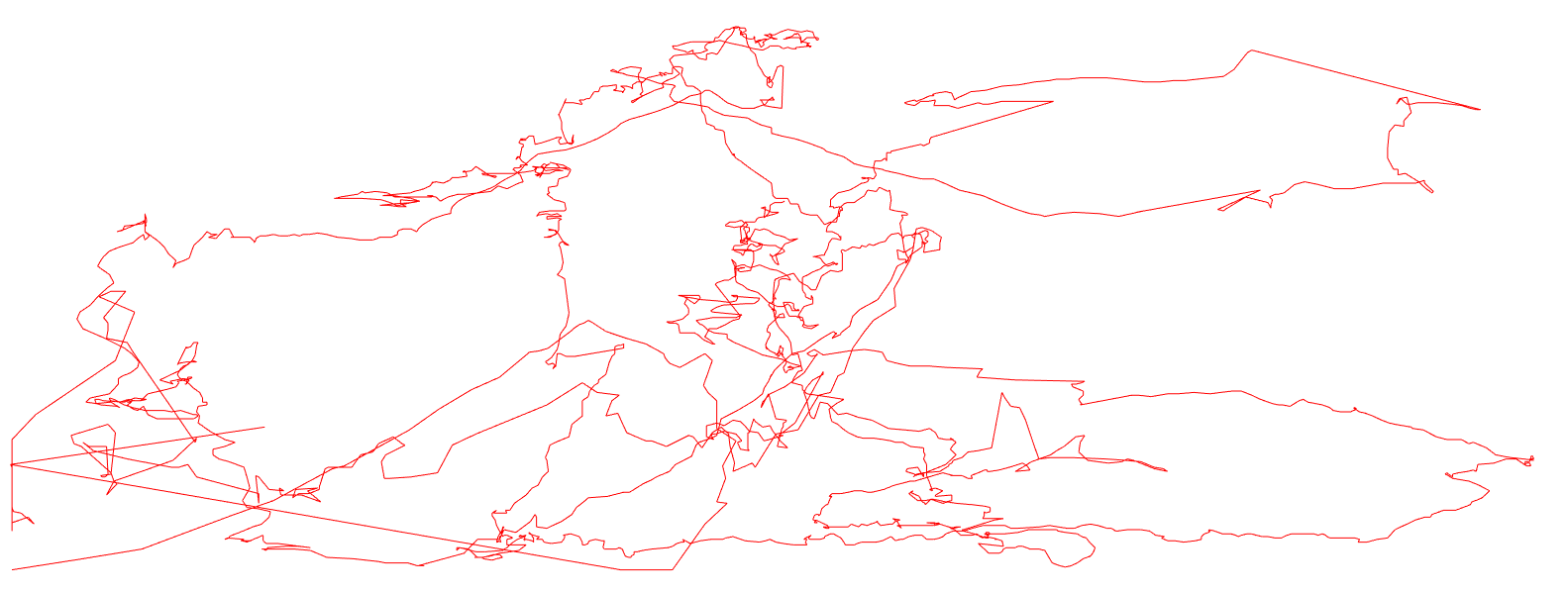}
		\caption{LaSOT Goldfish-4}
	\end{subfigure}
	\caption{Plots of annotations from human labelers}
	\label{fig:human_quality}
\end{figure}

Annotation of thousands of frames is a difficult job. Different labelers may perceive the same video or frame differently. Especially, when it comes to the edge of images, occlusion or partially visibility. Not only the differences between labelers vary a lot, the variation within one labeler in different frames is also considerably large. In Fig.\ref{fig:human_quality}, we pick some sequences which has fixed camera view and plot their human annotations. It's obvious that these trajectories are non-smooth, jumpy and have aliasing-like patterns. It can be even worse when the camera is also moving.

\subsection{Anomaly Frame Detection and Evaluation}
\label{sec:anomaly_detect_eval}
In this section, we uses the below proposed method in Sec.~\ref{sec:proposed_method} as a reference to detect outlier annotations and evaluate its performance.

\begin{figure}
	\begin{subfigure}[t]{.22\textwidth}
		\centering
		\includegraphics[width=\linewidth]{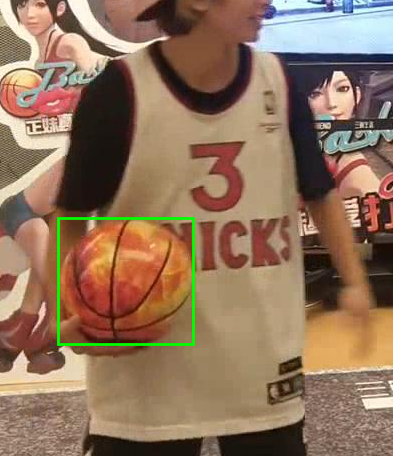}
		\caption{basketball-12 1st frame}
	\end{subfigure}
	\hfill
	\begin{subfigure}[t]{.22\textwidth}
		\centering
		\includegraphics[width=\linewidth]{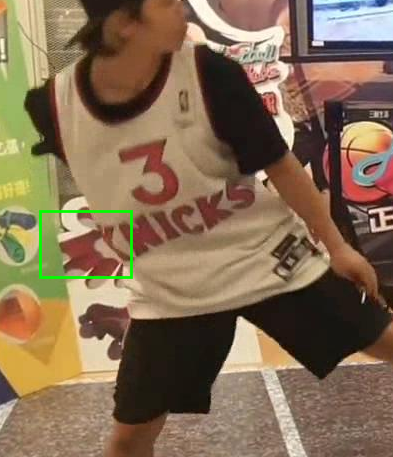}
		\caption{basketball-12 841st frame \textbf{object is behind the human}}
	\end{subfigure}

	\medskip

	\begin{subfigure}[t]{.22\textwidth}
		\centering
		\includegraphics[width=\linewidth]{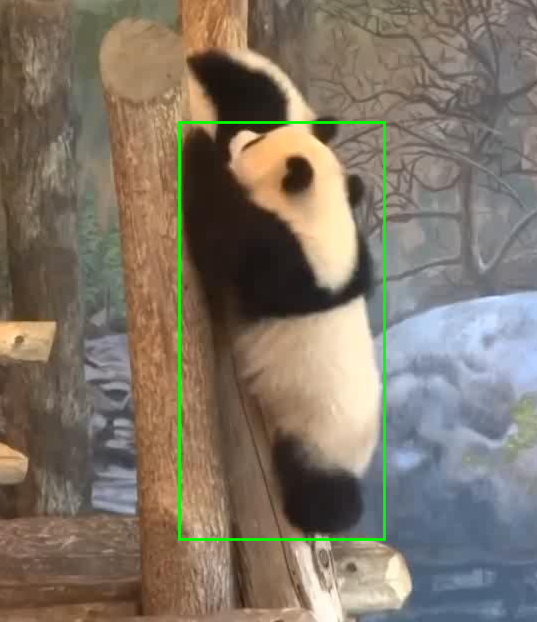}
		\caption{bear-15 1st frame}
	\end{subfigure}
	\hfill
	\begin{subfigure}[t]{.22\textwidth}
		\centering
		\includegraphics[width=\linewidth]{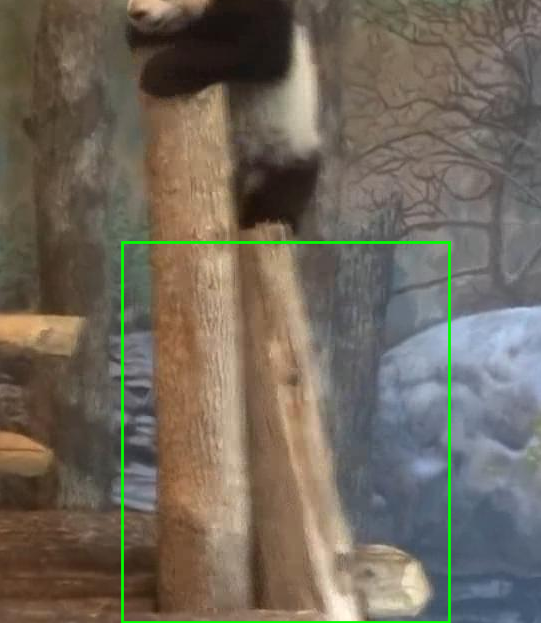}
		\caption{bear-15 851st frame \textbf{object is behind the tree}}
	\end{subfigure}

	\medskip

	\medskip

	\begin{subfigure}[t]{.22\textwidth}
		\centering
		\includegraphics[width=\linewidth]{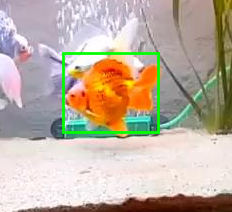}
		\caption{goldfish-18 1st frame}
	\end{subfigure}
	\hfill
	\begin{subfigure}[t]{.22\textwidth}
		\centering
		\includegraphics[width=\linewidth]{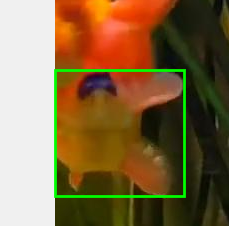}
		\caption{goldfish-18 1643rd frame \textbf{object has out-of-view rotation}}
	\end{subfigure}

	\caption{Examples of inaccurate annotations from LaSOT dataset}
	\label{fig:lasot_annotation}
\end{figure}

\begin{figure}
	\begin{subfigure}[t]{.22\textwidth}
		\centering
		\includegraphics[width=\linewidth]{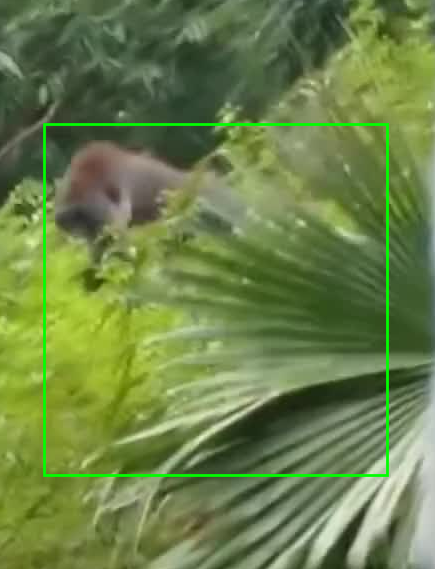}
		\caption{Train000002 1st frame}
	\end{subfigure}
	\hfill
	\begin{subfigure}[t]{.22\textwidth}
		\centering
		\includegraphics[width=\linewidth]{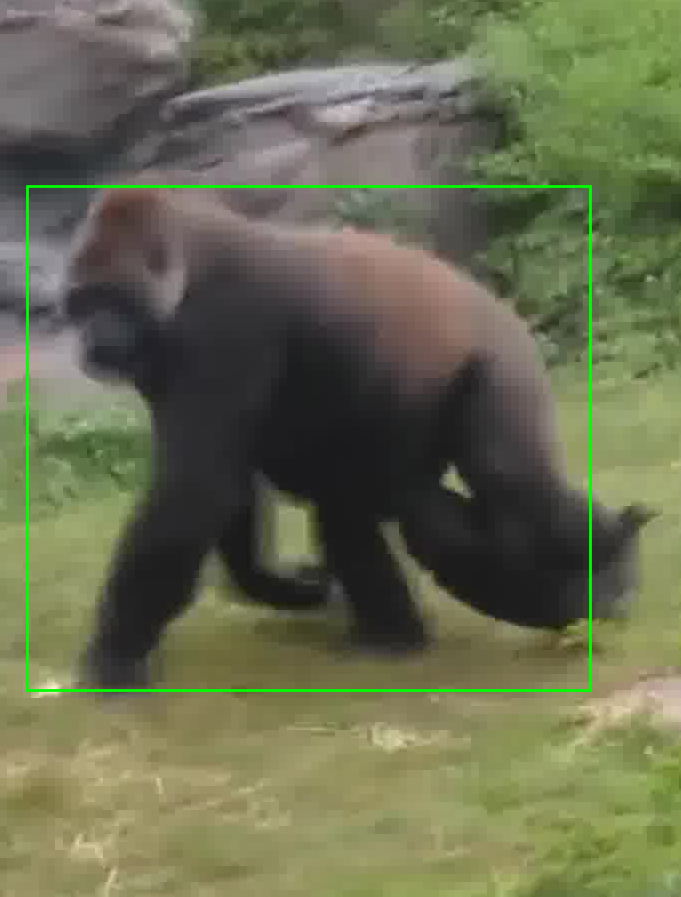}
		\caption{Train000002 25th frame \textbf{miss to include part of the foot}}
	\end{subfigure}

	\medskip

	\begin{subfigure}[t]{.22\textwidth}
		\centering
		\includegraphics[width=\linewidth]{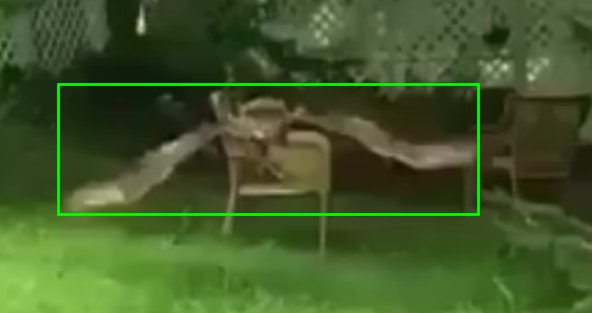}
		\caption{Train000178 1st frame}
	\end{subfigure}
	\hfill
	\begin{subfigure}[t]{.22\textwidth}
		\centering
		\includegraphics[width=\linewidth]{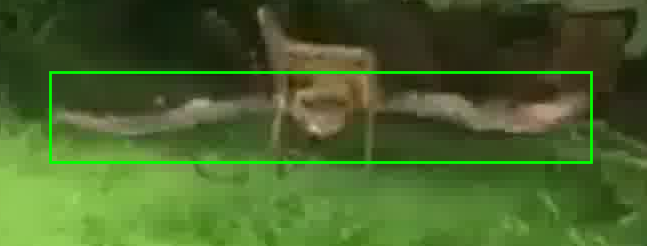}
		\caption{Train000178 23rd frame \textbf{include the chair legs}}
	\end{subfigure}

	\medskip

	\medskip

	\begin{subfigure}[t]{.22\textwidth}
		\centering
		\includegraphics[width=\linewidth]{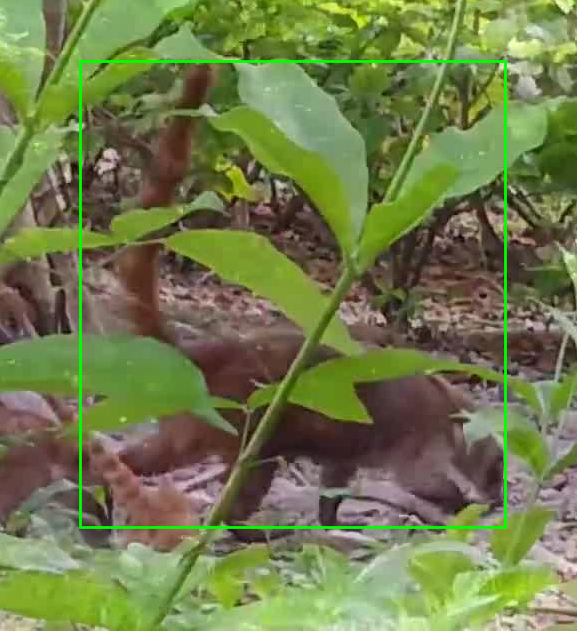}
		\caption{Train000284 1st frame}
	\end{subfigure}
	\hfill
	\begin{subfigure}[t]{.22\textwidth}
		\centering
		\includegraphics[width=\linewidth]{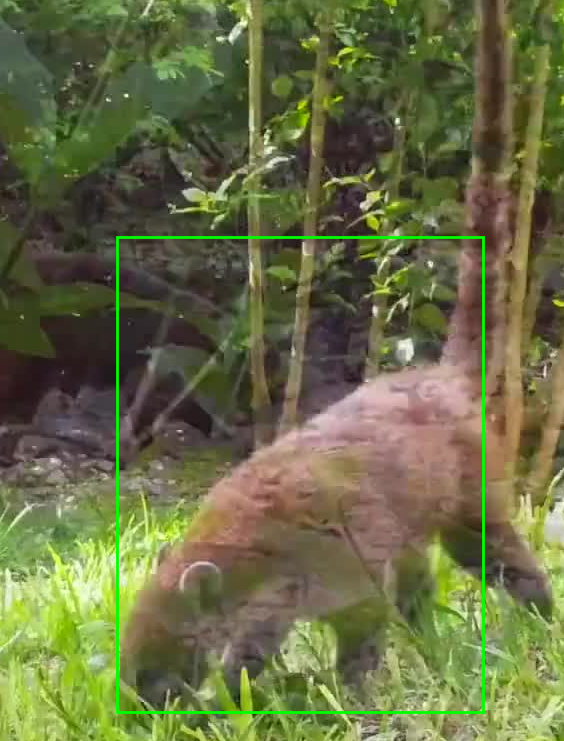}
		\caption{Train000284 5th frame \textbf{miss the tail and part of the leg}}
	\end{subfigure}

	\caption{Examples of inaccurate annotations from Got-10K dataset}
	\label{fig:got10k_annotation}
\end{figure}

After we obtain the smoothed trajectories and re-project it to its original plane, we can measure the distance between the annotated point and the smoothed point. If the distance is larger than a threshold $\tau$, it's a good candidate of an outlier. We can find human annotation errors in all of these datasets. Since the required quantity of annotation is normally huge and the quality of different labelers are different, it's very easy to let wrong annotations slip through.  In this example, we set the Euclidean distance threshold as $\tau = 100$ pixels, we then examine these outliers. A few example are extracted from LaSOT dataset and Got-10K dataset, shown in Fig.~\ref{fig:lasot_annotation} and Fig.~\ref{fig:got10k_annotation}.

There are a few major types of misalignment:
\begin{enumerate}
	\item \textbf{Partially or fully occlusion}. There are different strategies: only annotate what's visible; assume static size and infer its position; constant velocity interpolation and etc. In all cases, labelers have more freedom and introduce higher variances.

	\item \textbf{Close to boundary or out of view}. Labelers may skip the annotation or try to infer the location out of the view. In the case when object moves out of view, it's similar to the occlusion scenario, they may either only annotate what's visible or infer their position based on the static size assumption.

	\item \textbf{Object deformation}. When the tracked object has a large deformation due to out-of-plane rotation, it introduces freedom for labelers to interpret. The rule of thumb is to have a bounding rectangle that covers the whole object. But often times, the center of mass, the main torso, succinctness of the annotated box are also factoring in the bounding box decision.
\end{enumerate}

\begin{figure}
	\centering
	\includegraphics[width=\linewidth]{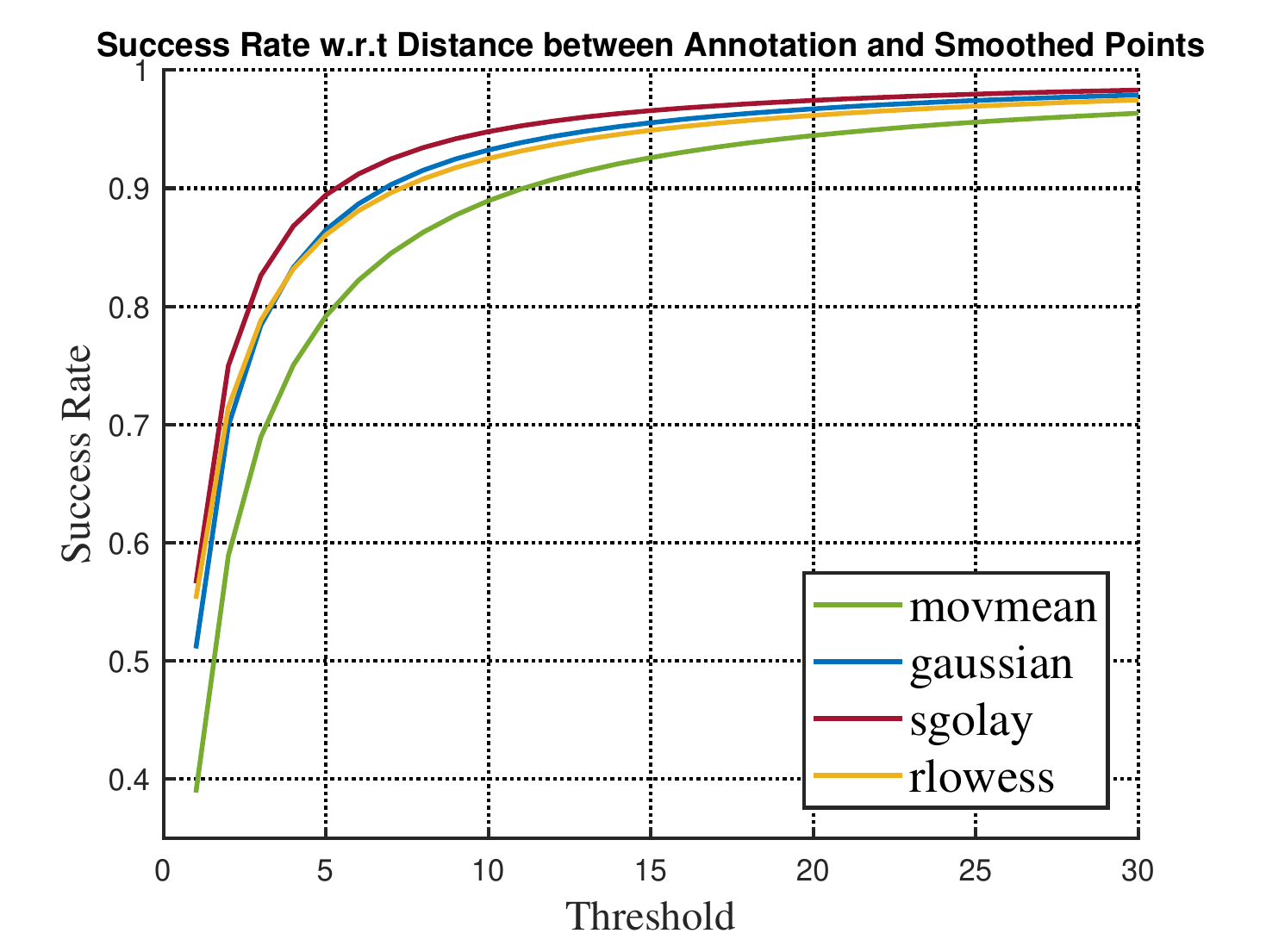}
	\caption{success rate at different threshold}
	\label{fig:success_rate}
\end{figure}

To give an overview of the general annotation quality, we plot the success rate with respect to the Euclidean distance threshold $\tau$ in Fig.~\ref{fig:success_rate}. If the distance between annotation center and the smoothed point is smaller than $\tau$, then the frame is considered successful. Note, neither the smoothed trajectories nor the annotation is perfect, so this analysis is a proof-of-concept. We can see when $\tau > 12$, the success rates are all above 90\%. It shows the annotation is of high quality with a good align rate with smoothed version, but it does show there are outliers. Also note, because the smoothed trajectory is not very well at predicting sudden motion change or back-and-forth pattern, in those type of video sequences, the failure is mostly due to the smooth technique itself.

\section{Proposed Method}
\label{sec:proposed_method}
In this section, we explain our proposed method to align the human annotation better with the real ground truth.

Many video sequences are captured from hand-held cameras, there are inevitably camera motion patterns. To make the target object visible in the frame, they often move to follow the target. Thus we cannot always assume the images are static and containing target object motion only.

\subsection{Image Alignment}
\label{sec:image_alignment}
We need to register the image sequences into a same coordination system, so that from their positions, it should form a smooth trajectory. To compensate for the camera motion, we can utilize the image alignment technology.

A generic image alignment algorithm compares two consecutive images by extracting their points of interest, through SIFT, HoG, ORB, SURF etc. and matches these points into pairs. Once these points are matched, a homography is inferred to recover the transformation relationship between the two images, for example using RANSAC.

The challenges we are facing in a video tracking scene are two-folds: 1) The view range is constantly changing throughout the video. We need the ability to locate the object in a fixed coordination system with a non-static background. In contrast, in two images alignment scenario, only common region of interest is considered. 2) Although we always assume a reasonable frame rate, the image blurry still present when either the object is in a fast motion or the camera motion is fast. In this case, keypoint techniques will fail.

\begin{figure}
	\centering
	\includegraphics[width=\linewidth]{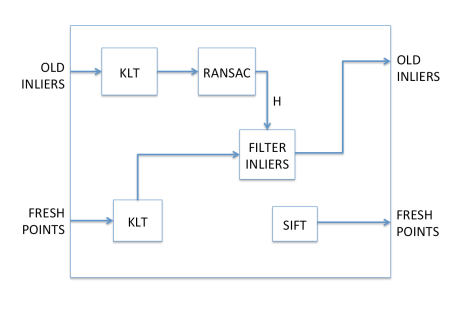}
	\caption{Illustration of the adaptive update strategy in RSRT~\cite{dutta2014video}}
	\label{fig:rsrt}
\end{figure}

In ~\cite{dutta2014video}, Dutta et al. proposed a simple update strategy Robust staged RANSAC Tracking (RSRT) to keep the connection of consecutive images while adapt to the motion change. They estimate the homography using the inlier points from the two previous image. Then the keypoints extracted from the previous image are tested against the homography. The passed test keypoints will become the new inliers into the next frame. The idea is illustrated in Fig.~\ref{fig:rsrt}. This could solve most of the alignment problem, but when the image is blurry due to the fast motion, it's hard to extract enough keypoints, and it's always impossible to utilize only a few good matched keypoints to do a reasonable homography inference.

Another classic alignment algorithm is enhanced correlation coefficient maximization (ECC)~\cite{evangelidis2008parametric}. This is a template based algorithm, it tries to find a pixel level alignment using an iterative method to approximate a non-linear function. In the case of blurry image, the assumption is the correct object region will still yield the local minimum, although the overall matching score is worse than when the image is clear. In this way, even when there is not enough keypoints, we can still have a good estimation of the homography of the next frame. The problem with ECC is due to its nature of template based, it can only apply to the target object area and is not suitable for a full image alignment, and it's more expensive than a keypoint based algorithm. Thus, it is only served as a secondary method to counter the blur situation.

So our image alignment algorithm is as follows:
\begin{itemize}
	\item Extract keypoint and use RSRT~\cite{dutta2014video} strategy to construct homography $H_i$ and update the status.
	\item In the case where keypoints are less than the threshold ($\tau = 20$ in our experiment), we fall back to use ECC where the template is extracted from the previous target object area. The result is also the homography $H_i$. But we don't update the keypoints when using ECC.
	\item When we can extract enough keypoints in the frame, we will resume the RSRT strategy and continue.
	\item When both RSRT and ECC cannot achieve a good estimation, we mark the trajectory as failed, because we are not confident that the extracted points are of good quality. This is one limitation of this approach.
\end{itemize}

After we extract the homography of each frame and convert the annotated ground truth to its first image coordination, we now have a trajectory that is invariant to the camera motion and it can be calibrated from a canonical view.

\subsection{Trajectory Smoothing}
\label{sec:trajectory_smoothing}
Trajectory smoothing is widely used in signal processing, finance, robotics and many other areas. Although the scenarios are significantly different, the core ideas are the same. It uses prior knowledge and assumption to rectify the point series and reduce noises. In our tracking scenario, we want to use this method to reduce the noise introduced by the human errors, as shown in Fig.~\ref{fig:human_quality}. We assume the moving object should exhibit a smooth, continuous and predictable motion, rather than an abrupt, jittering and unpredictable one.

In this paper, we try a few common ones and experiment to see what's the major factors that affect the quality.

\begin{itemize}
	\item Moving Average is the most common way to smooth the trajectory. It smooths the trajectory based on a local average of a few points.
	\item Gaussian smoothing is one type of kernel smoothing. It uses a Gaussian kernel to assign weights at each convolution step.
	\item Savitzky–Golay smoothing~\cite{savitzky1964smoothing} is trying to fit higher order moments by using polynomial smoothing terms.
	\item Local regression is a non-parametric method using nearby points to fit the curve without a predefined model as the other smoothing operators. One classic approach is Lowess (Locally Weighted Scatter-plot Smoother) ~\cite{cleveland1981lowess}.
\end{itemize}

We apply different windows and settings to validate which better serve the purpose of getting a well-aligned trajectories.

Once the trajectory is smoothed, we will convert the bounding box back to its original image coordination system using the previously computed homography from Sec.\ref{sec:image_alignment} .

\section{Experiment}
\label{sec:experiment}
\subsection{Smoothed Data as Better Ground Truth}
\label{sec:smoothed_data_better_ground_truth}
\begin{figure}
	\begin{subfigure}[t]{.22\textwidth}
		\centering
		\includegraphics[width=\linewidth]{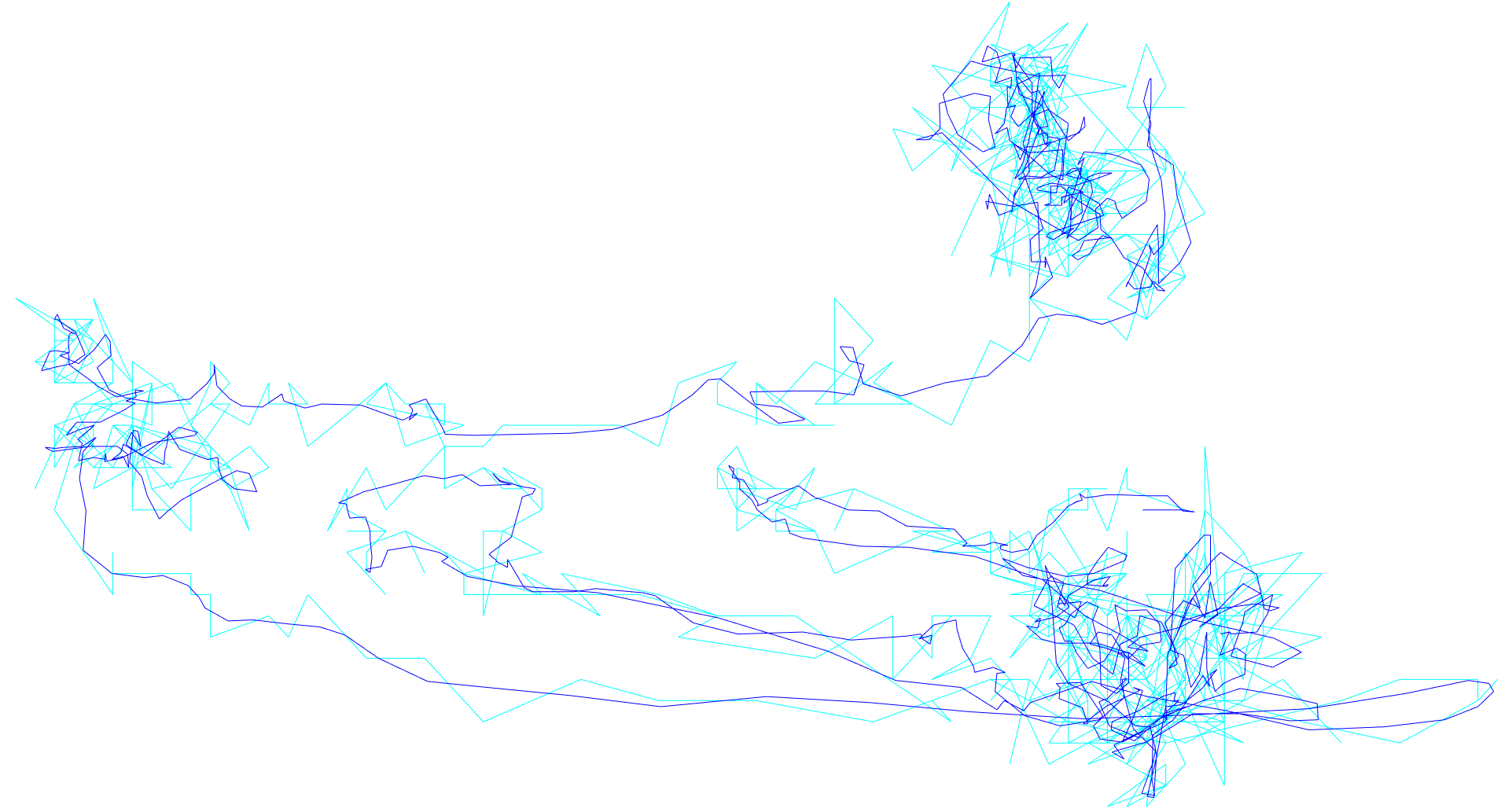}
		\caption{TB-50 Biker}
	\end{subfigure}
	\hfill
	\begin{subfigure}[t]{.22\textwidth}
		\centering
		\includegraphics[width=\linewidth]{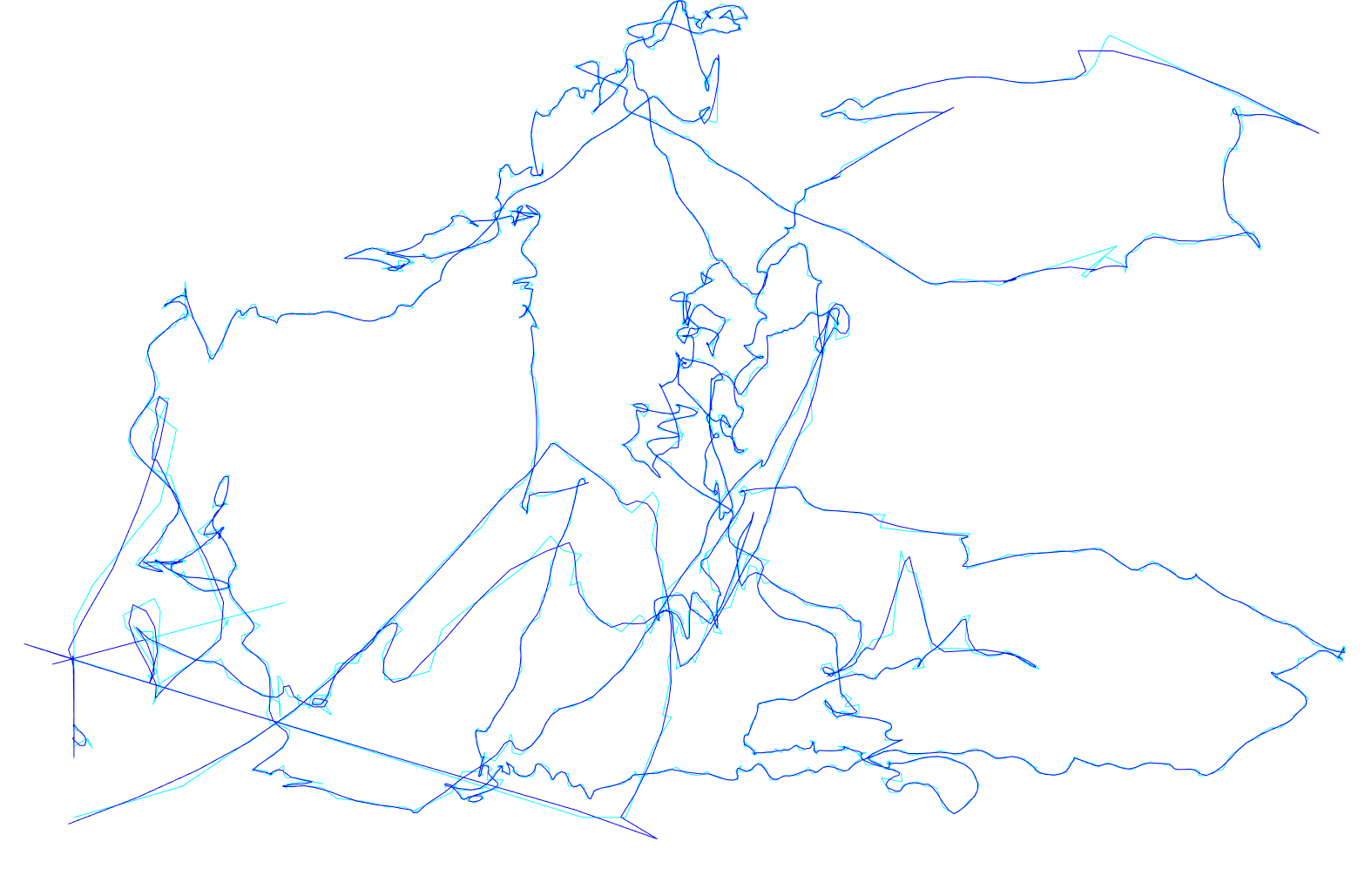}
		\caption{LaSOT Goldfish-4}
	\end{subfigure}

	\caption{Smoothed trajectory compared with its original annotations (better visualized when zooming-in and color version)}
	\label{fig:smoothed}
\end{figure}

Applying the smoothing techniques as explained in Sec.~\ref{sec:proposed_method}, and after converting the trajectory back, we obtain the final smoothed annotations. We take some examples as shown in Fig.~\ref{fig:smoothed}. The blue curve shows the smoothed trajectory, while the cyan one is the original annotation. We can see clearly the smoothed one removes the jittering effects and yields a more consistent trajectory.

\subsection{Smoothed Data for Training}
\label{sec:smoothed_data_training}
\begin{table*}
	\centering
	\caption{Retrained model statistics using different smoothed ground truth}
	\begin{tabular}{|c|c|c|c|c|c|c|}
		\hline
		& & baseline & $\tau=30$ & $\tau=20$ & $\tau=10$ & $\tau=5$ \\
		\hline
		\multirow{2}{*}{movmean} & replaced \%	& 0\% & 3.66\% &  \cellcolor{green} 5.54\% & 11.07\% & 20.83\% \\
		& OPE Precision	& 0.378 & 0.379 & \cellcolor{green} 0.383 & 0.364 & 0.351  \\
		\hline
		\multirow{2}{*}{Gaussian} & replaced \%	& 0\% & 2.12\% & 3.29\% & \cellcolor{green} 6.77\% & 13.53\% \\
		& OPE Precision	& 0.378 & 0.378 & 0.379 & \cellcolor{green} 0.387 & 0.362 \\
		\hline
		\multirow{2}{*}{Savitzky-Golay} & replaced \%	& 0\% & 1.70\% & 2.57\% & \cellcolor{green} 5.22\% & 10.58\% \\
		& OPE Precision	& 0.378 & 0.378 & 0.379 & \cellcolor{green} 0.388 & 0.365  \\
		\hline
		\multirow{2}{*}{Lowess} & replaced \%	& 0\% & 2.54\% & 3.84\% & \cellcolor{green} 7.49\% & 13.96\% \\
		& OPE Precision	& 0.378 & 0.378 & 0.379 & \cellcolor{green} 0.382 & 0.362 \\
		\hline
	\end{tabular}
	\label{tab:result}
\end{table*}

After generating the smoothed trajectory, we can also use them to get a better training data. To test our hypothesis, we can put thresholds on how much we trust the original labeled data. We will replace the training sample if the difference between the original label and the smoothed data point is greater than a threshold $\tau$. The extreme case would be fully replaced by the smoothed trajectories. As an example from Fig.~\ref{fig:success_rate}, using Savitzky-Golay method, if we set $\tau=5$ pixels, meaning if the Euclidean distance is larger than 5 pixels, we will replace the human annotated point with the smoothed point. In this particular example, around $10.58\%$ datapoints will be replaced by the smoothed version, see Table.~\ref{tab:result} for reference. To replace the training sample, we use the smoothed data point center location and use the original annotated object size as the new training patch.

\begin{figure}
	\centering
	\includegraphics[width=\linewidth]{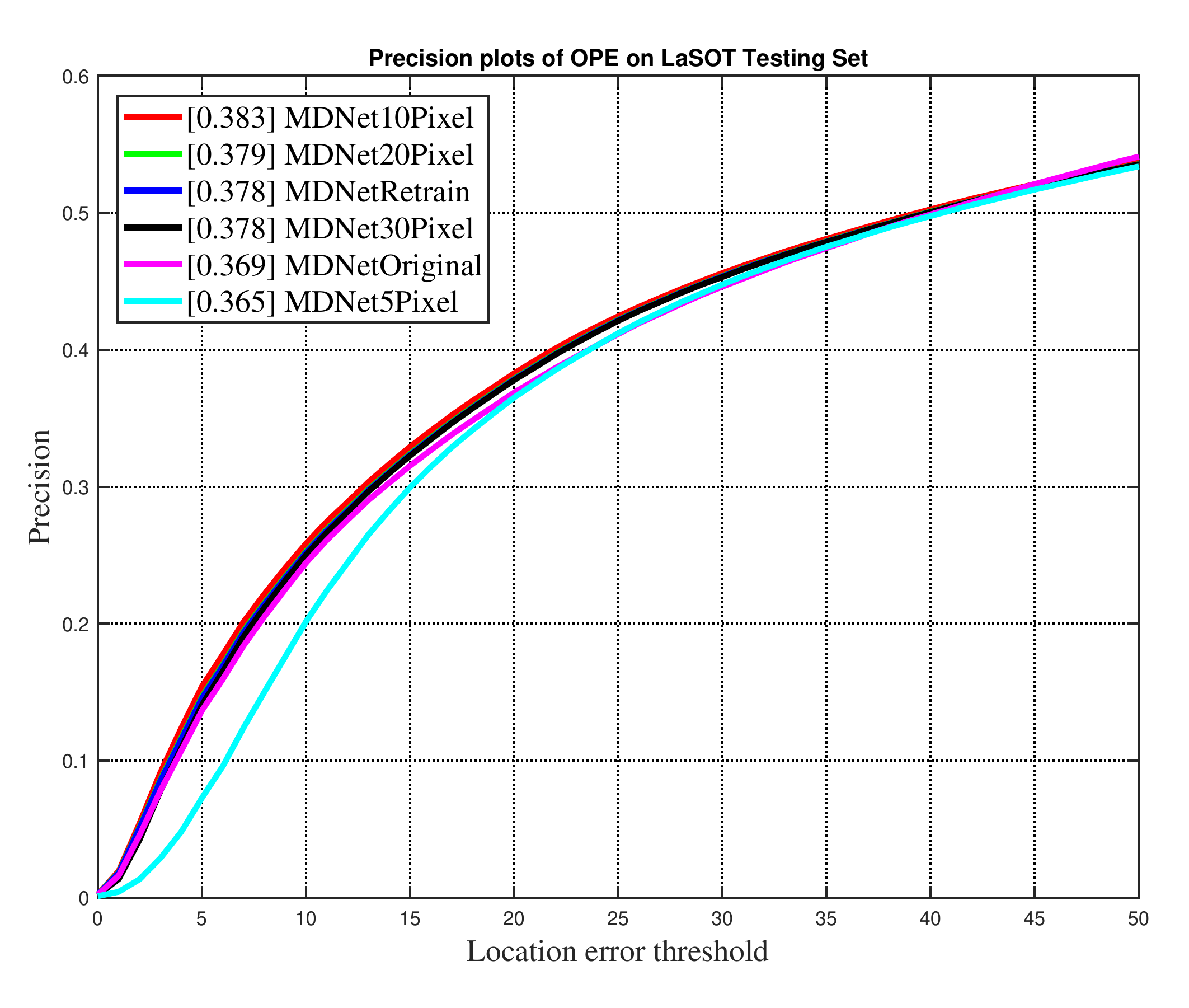}
	\caption{Compare the retrained evaluation OPE precision at different threshold using Savitzky-Golay smoothing technique}
	\label{fig:retrain_ope}
\end{figure}

We use the pre-defined training/testing split from LaSOT dataset~\cite{fan2019lasot}. We apply the patch replacement in the training phase, and then we use the original annotation as the ground truth to evaluate the OPE precision. We retrained the model 10 times and use the average as the final reported results.
The result is shown in Table.~\ref{tab:result} and one OPE precision plot example is shown in Fig.~\ref{fig:retrain_ope}.

We can see from the result, if the replaced points are very small, it has no effects to the final result. This is expected as when the distribution of the training data doesn't change much, we won't expect any changes. As the replaced points gets to around 3\% of the total training set, we start seeing promising improvement. As the replaced points gets to around 5\%, we see a meaningful improvement in terms of the precision. However, as the size of replaced points increases, we start seeing the performance decreases, once it reaches more than 10\%, the result is already below the baseline.
This can be explained by the fact that the smoothed trajectory does change the target object appearance. Using it as a guideline to correct outliers does make sense, but if used too much, it will affect the effectiveness of the model training and the model may not learn well.

Reading the "replaced \%" in Table.~\ref{tab:result}, we can see the distribution w.r.t. the $\tau$. For example, movmean changes the original trajectory more dramatically than others, it has a consistent higher replaced percentage than other methods, and when $\tau=5$, it replaces more than 20\% of the points. Due to this, the movmean optimal threshold occurs earlier than other smoothing techniques. This also shows the performance of training data correlates more to the replaced percentage than the threshold itself.
It's also interesting to see that Gaussian, Savitzky-Golay and Lowess methods are exhibiting similar behavior. It's a sign that for training purpose, the smoothing techniques themselves may not matter much, as long as we find the proper replaced percentage.

\section{Discussion}
\label{sec:discussion}
This strategy besides correcting the annotation and training data, can also be used early on to guide the annotation along the way and help reduce the labeling workload. The annotation can be applied in a first round on key frames. Then we can use this method to smooth the whole trajectory and pinpoint the center location of where the object should be at in each frame. During the second round, each frame can be annotated around the referenced point. This is especially useful when the object is occluded or  partially visible, otherwise human labelers would have a hard time to decide the mass center w.r.t. the original object.

This method also has a few limitations:

\begin{itemize}
	\item It relies on the image alignment technology, if the alignment fails, we won't be able to get a meaningful trajectory after that point. In that way, the sequence cannot be optimized.
	\item The corrected annotation, without a second round of adjustment, may not accurately bound the object well, because this method does not change the bounding box size.
	\item If the object presents a back and forth motion pattern, the smoothing technology cannot capture it well. It always tries to bring the control points closer together and make the smoothed trajectory off its true ground truth. For those sequences, the smoothed version is normally worse than the original annotations.
\end{itemize}

\section{Conclusion}
\label{sec:conclusion}
In this paper, we propose a framework to help improve the human annotation for video tracking sequences. We can use our robust image alignment technique to register the frames into a canonical view. Then we apply the smoothing algorithms to smooth the annotated results. Once the smoothed trajectory is re-projected to their original image, it can be used as a guideline to correct the human annotation results. It has several applications, it can help correct annotation outliers. With a certain amount of corrections, it can also help improving the training data quality and improve the models. Potentially, it can extrapolate the missing annotation frames, this would speed up the annotation process. During experiment, we show the trajectory is much better after the smoothing. And we can use the smoothed data to re-trained models. The results show different smoothing techniques are similar from a training model perspective, but overall trend shows, by using the right amount of correction, they all can improve the performance.

{\small
\bibliographystyle{unsrt}
\bibliography{improve}
}

\end{document}